\documentclass{tlp}
\usepackage{comment}
\usepackage{todonotes}
\usepackage[semibold,scale=0.875]{sourcecodepro}
\usepackage[utf8]{inputenc}
\usepackage{amsmath}
\usepackage{graphicx}
\usepackage{multirow}
\usepackage{amssymb}
\usepackage{microtype}
\usepackage{url}
\usepackage{xcolor}
\usepackage{listings}
\let\proof\relax

\usepackage{amsthm}
\usepackage{xspace}
\usepackage{booktabs}
\usepackage[T1]{fontenc}

\usepackage{cleveref}
\def\beq{\begin{equation}}
\def\eeq#1{\label{#1}\end{equation}}
\def\ba{\begin{array}}
\def\ea{\end{array}}

\def\paramset#1{{#1}^\Downarrow}

\newcommand{\HH}{\mathcal{H}}

\newcommand{\boldX}{\mathbf{X}}

\newcommand{\boldt}{\mathbf{t}}
\newcommand{\boldr}{\mathbf{r}}

\newcommand{\boldu}{\mathbf{u}}

\crefformat{footnote}{#2\footnotemark[#1]#3}

\def\property{{\em property}\xspace}

\DeclareRobustCommand{\modelsht}{%
  \mathrel{\mathchoice
   {{\models}_{\!\!\!\!\scriptscriptstyle\rm ht}}
   {{\models}_{\!\!\!\!\scriptscriptstyle\rm ht}}
   {{\models}_{\!\!\!\!\scriptscriptstyle\rm ht}}
   {{\models}_{\!\!\!\!\scriptscriptstyle\rm ht}}
}}
\newcommand{\numeral}[1]{\overline{#1}}

\newcommand{\tuple}[1]{\ensuremath{\langle #1 \rangle}}

\newcommand{\trdef}[3]{\big\langle#1,#2,#3\big\rangle}
\newcommand{\ph}{\mathit{PH}}

\def\definition{module\xspace}
\def\definitions{modules\xspace}
\def\clingo{{\it clingo}\xspace}

\def\mp{\mathcal{P}}

\newcommand{\eqdef}%
{%
	\mathrel{\vbox{\offinterlineskip\ialign%
	{%
		\hfil##\hfil\cr%
		$\scriptscriptstyle\mathrm{def}$\cr%
		\noalign{\kern1pt}%
		$=$\cr%
		\noalign{\kern-0.1pt}%
	}}}%
}

\newcommand{\pdef}{{\lambda}}

\newcommand{\EM}[1]{\mathit{EM}(#1)}

\newcommand{\At}[1]{\ensuremath{\mathcal{A}}_{#1}}

\newcommand{\PH}{\mathit{PH}}

\def\rar{\rightarrow}
\def\lrar{\leftrightarrow}

\newcommand{\G}{\mathfrak{G}}

\newcommand{\Module}[2]{\ensuremath{\tuple{{#1},{#2}}}}

\lstdefinelanguage{anthem}
{
	keywords={and, exists, forall, integer, not, or},
	otherkeywords={assume, axiom, backward, forward, input, lemma, output, spec},
	morekeywords={},
	morecomment=[l]{\#},
	sensitive,
}

\def\clingo{{\em clingo}\xspace}

\definecolor{gray}{rgb}{0.33,0.33,0.33}
\renewcommand*\copyright{{%
  \usefont{T1}{lmr}{m}{n}\textcopyright}}
\newtheorem{theorem}{Theorem}

\newtheorem{proposition}[theorem]{Proposition}

\begin{document}

\jnlPage{1}{8}
\jnlDoiYr{2021}
\doival{10.1017/xxxxx}

\title[Parametric Modular Answer Set Programs Made Declarative]
{Parametric Modular Answer Set Programs\\
Made Declarative}



%


\begin{authgrp}
\author{\sn{Jorge} \gn{Fandinno}}
\affiliation{University of Nebraska Omaha, USA}
\author{\sn{Yuliya} \gn{Lierler}}
\affiliation{University of Nebraska Omaha, USA}
\author{\sn{Torsten} \gn{Schaub}}
\affiliation{University of Potsdam, Germany}
\end{authgrp}

\history{\sub{xx xx xxxx;} \rev{xx xx xxxx;} \acc{xx xx xxxx}}

\maketitle


\begin{abstract}
    In this paper, we explore the concept of modularity in first\nobreakdash-order answer set programming (ASP).
    We introduce a new formalism called parametric modular logic programs, which allows defining subprograms with
    parameters and intensionality statements.
    We demonstrate how this formalism can capture the semantics of clingo-programs with \emph{collective control},
    a feature that enables structuring and instantiating subprograms.
    We provide theoretical foundations for modular ASP, illustrate its usefulness,
    and connect to traditional non-modular ASP.
\end{abstract}


\section{Introduction}
Answer set programming (ASP)
is a well-established declarative logic programming paradigm under the \emph{answer set/stable models semantics}~\citep{gellif88b}.
The ASP methodology relies on devising a logic program so that its answer sets are in one-to-one correspondence to the solutions of the target problem.
This approach is fully declarative, since the logic program only describes a problem and conditions on its solutions, not the way to obtain them. The latter is  delegated to systems called answer set solvers.
A~software engineer is then tasked with translating problem's specifications into the formal language of logic programs rather than devising algorithms for obtaining solutions to a problem.
Under these circumstances both the art of programming and the construction of an argument for code's correctness become more direct~\citep{cafali20a,falilusc20a,hanlir25a,fahaliglheschstli25a}.
%
This is essential in building trustworthy software solutions, making ASP a viable role player in the current programming landscape.
Yet, to realize the full potential of ASP more attention to its theoretical and practical aspects have to be paid.
Here, we target
advancing theoretical foundations of the ``modularity''  frontiers of this programming paradigm.

Modularity is one of the key techniques in principled software development, and it is essential for modeling large-scale practical applications.
It provides a solid abstraction to study distinct approaches to solving a problem by identifying modules/parts of code that are responsible for different aspects and helps in constructing arguments of correctness by decomposing the argument about the whole program into arguments about its components~\citep{cafali20a}.
In general, traditional ASP programs lack modularity because subsections of the code cannot be evaluated in isolation.
Most well known variants of ASP systems\,---\,{\em lparse-smodels} pair, earlier versions of \clingo, {\em dlv}\,---\,expect a program at its full length bypassing any support for modularity.
Yet, researchers realized early the importance of this aspect. Theoretical and practical advances exist.
For instance, \citet{oik06,oik08} and~\cite{jaoitowo07a} devise a propositional modular framework for answer set programming and introduce concepts of equivalence for such programs;
\cite{harlie16a} introduce first-order modular logic programs.
\citet{lif94e} proposed  {\em splitting} as a tool to ``modularize'' otherwise monolithic logic programs. Mentioned works were at the origin of more theoretical advances refining their proposals.
On the practical side,
\cite{gekakasc14a} describe the ${\mathit{ASP}\!+\!\mathit{control}}$ framework that, for example, allows encodings which include time horizons and assume repetitive code for each time step mimicking settings of planning domains.
This framework was implemented within \clingo~4.
A PDDL planner
{\sc plasp} backed up by ASP technology utilizes this feature~\citep{digelurosc17a}.
Multi-shot solving~\citep{gekakasc17a} became one of the staples of \clingo~5.
It prominently positions an interactive mode of programming with ASP making control over ASP ``modules'' simple.
%
Yet, control equates with a procedure that encroaches on the declarativness of ASP.
In this paper, we study how ``first-order modules'' and ``control'' can be given a declarative meaning, by tackling on particular kind of control that we call {\em collective}.
This work is a foundational step towards making the concepts pertaining (i)~ASP modules as they are used in practice and (ii)~control precise, well understood and~{\em declarative}.
This allow us to construct arguments of correctness for logic programs with collective control.

This paper is organized as follows.
We start by introducing a motivating example.
We then introduce theoretical preliminaries. Modular programs are introduced as the basis for what we call parametrized modular logic programs. We conclude by illustrating how parametrized modular programs capture the semantics of programs such as the one discussed as a  motivating example. We use this example to showcase possible formal claims that can be seen as arguments of correctness for modular ASP programs.

\section{Motivating Examples}\label{sec:motiv}

We start by introducing a simple example that illustrates the key concepts of parametric modular logic programs and collective control.%
\begin{lstlisting}[
    caption={\vspace{3pt}The {\em property} parametric-program}\label{lst:property},
    language=Prolog,
    numbers=right,basicstyle=\small\ttfamily,numbersep=-8pt,frame=single,
    aboveskip=0pt, belowskip=0pt
    ]
 #program base.
 q(0,0).
 #program property(k).
 q(N,k+1) :- q(N-1,k).
\end{lstlisting}%
Listing~\ref{lst:property} contains two subprograms {\tt base} and {\tt property(k)}.
Keyword {\tt base} marks
  a dedicated subprogram (with an empty parameter list).
 The {\tt base} subprogram in  Listing~\ref{lst:property}
 includes the single fact {\tt q(0,0)}.
 Without further control instructions, {\em clingo}  grounds and solves the {\tt base} subprogram only.
 This yields the standard behavior of answer set systems. In order to process the other subprograms such as {\tt property(k)},  control instructions have to be explicitly stated.
For instance, assume the \property parametric-program
in the context of  the following \emph{control}:
\begin{equation}
  \tag{A}\label{eq:A}
  \parbox{\dimexpr\linewidth-3.5em}{%
    \strut
{\em given a positive integer $n$, compute the answer sets of a program constructed from the \property program
by appending the content of its {\tt base} program fact $q(0,0)$ to $n$ copies of the rules in {\tt property(k)} where parameter~$k$ ranges from 0 to $n-1$.}
    \strut
  }\hspace{1em}
  \end{equation}
The python routine
presented in Listing~\ref{lst:controlA} below implements control~\eqref{eq:A}.
\begin{lstlisting}[caption={\vspace{3pt}The implementation of control~\eqref{eq:A}}\label{lst:controlA},numbers=right, firstnumber=5,language=Python,numbersep=-10pt,frame=single,basicstyle=\small\ttfamily,float]
#script(python)
from clingo.symbol import Number
def main(prg):
  n = prg.get_const("n").number
  subs = [("base",[])]
  for k in range(0,n):
    subs.append(("property",[Number(k)]))
  prg.ground(subs)
  prg.solve()
#end.
 \end{lstlisting}
If the contents of Listings~\ref{lst:property} and~\ref{lst:controlA} are stored in a file named {\tt property.lp} then the command line {\tt clingo
property.lp -c n=100} (the presented code is compatible with \clingo version 5.6) returns the unique answer set
 $\{q(0,0)\;\;\ q(1,1)\;\;\ q(2,2)\;\;\ \dots\;\;\ q(100,100)\}$.
Directive  {\tt -c n=100} provides a specific value 100 to be utilized in place of~{\tt n} by \clingo.
%
%
The {\bf while}-loop of the routine in Listing~\ref{lst:controlA}
appends to the base program~$n$ pairs of the subprogram named {\em property} where the placeholder~$k$ is replaced by numbers $0$ through~$n-1$.
In line 12, the instruction is given to ground all rules in~{\tt base} together with all rules in~the {\tt property} subprograms with their parameters~$k$ substituted by the corresponding integers.
In line 13, the resulting ground program is solved.
Thus,
the \property parametric-program together with the specified control is understood as the following ``typical'' logic program:
\begin{equation}
  \begin{aligned}
  &q(0,0).
  \qquad&& q(N,0+1) \texttt{ :- } q(N-1,0).
  \qquad&& q(N,1+1) \texttt{ :- } q(N-1,1).
  \\
  &\cdots
  &&q(N,99+1) \texttt{ :- } q(N-1,99).
  \end{aligned}  \label{lst:unfolded}
\end{equation}
Though extremely simple, the program in Listing~\ref{lst:unfolded} illustrates the key concepts of parametric modular logic programs and collective control that appear in planning problems.

A parametric program together with its control can be seen as the means to concisely described what can be a lengthy program. Furthermore, it naturally groups rules of a program together. In the closing section, we illustrate how such grouping can be used in constructing arguments about formal properties of a program augmented with control based on the properties about its individual subcomponents.
Let us summarize what we call a {\em collective  control} exemplified  by~\eqref{eq:A}: this control
\begin{itemize}
\item collects   the subprograms of interest with the values for the respective parameters;
\item instantiates the respective parameters by their values in each of the subprograms;
\item grounds the collection as if it were the union of rules;
\item solves the ground program.
\end{itemize}
In what follows we provide the declarative semantics to \clingo programs with the collective control.
The proposed  semantics allow us to associate meaning with each individual component of a
program with control without referring to concatenation and grounding.

%

\section{Theoretical Preliminaries}
We open this section by describing syntax and semantics of normal logic programs with arithmetic~\citep{falilusc20a,lifschitz21a,falite24a}.
Then, we review the key concepts stemming from the multi-shot framework~\citep{gekakasc17a}.
%

\subsection{Logic Programs and their Semantics}\label{sec:lpsems}

\paragraph{Many-sorted first-order formulas.} 

We assume a many-sorted signature consisting of two sorts, \emph{integer} and \emph{general}, where the integer sort is a subsort of the general sort.
This signature contains arithmetic functions such as~$+$, $-$, and $\times$, whose argument and value sorts are integer, and comparison predicate constants such as~$\leq$ or~$>$, whose argument sorts are of sort general.
%
%
We use infix notation for these arithmetic function and predicate constants.
This signature also contains as object constants of sort integer all numerals~$\overline{0},\overline{1},\overline{-1},\overline{2},\dots$.
We often identify numerals with integers and just write~$0$ instead of~$\overline{0}$, but sometimes it is important to distinguish them.
When it is important to distinguish a numeral that is an element of the signature from an underlying integer we will use an overline.
This notation allows us to distinguish between expressions $\overline{5+3}$ and $\overline{5}+\overline{3}$. The former is identified with the numeral $\overline{8}$, whereas the later is identified with an algebraic expression that encodes function application to two numerals $\overline{5}$ and $\overline{3}$.
In the remainder we use the overline only in cases exemplified by expression~$\overline{5+3}$.
We assume that for every sort, an infinite sequence of \emph{object
variables} of that sort is chosen.
\emph{Terms} over a signature~$\sigma$ are defined recursively as usual with the addition that arguments of functions
have to be of appropriate sorts.
\emph{Atomic formulas} over~$\sigma$ are either (i)~expressions of the form $p(t_1,\dots,t_n)$, where~$p$ is a predicate
constant and $t_1,\dots,t_n$ are terms such that their sorts are
subsorts of the argument sorts~$s_1,\dots,s_n$ of~$p$, or (ii)~expressions of the form $t_1=t_2$, where~$t_1$ and~$t_2$ are terms.
%
\emph{(First-order) formulas} over signature~$\sigma$ are formed from atomic formulas
and the 0-place connective~$\bot$ (falsity) using the binary
connectives $\land$, $\lor$, $\to$ and the quantifiers $\forall$, $\exists$.
The other connectives are treated as abbreviations: $\neg F$ stands for
$F\to\bot$ and~$F\lrar G$ stands for $(F\to G)\land (G\to F)$.
A term or other expression is called \emph{ground} if it contains no variables.
A ground expression is called \emph{precomputed} if it contains no arithmetic function constants.
For instance, $p(f(1))$ is precomputed, whereas $p(1+2)$ is ground but not precomputed.

A \emph{rule} is a formula of the form~$\mathit{Head} \leftarrow \mathit{Body}$ (or, $\texttt{Head\,:-\,Body}$ in source code format), where~$\mathit{Head}$ is an atomic formula or~$\bot$, and~$\mathit{Body}$ is a list of \emph{literals}, that is, atomic formulas possibly preceded by one or two occurrences of~$not$.
A \emph{program} is a set of rules.

Every rule can be identified with some first-order formula over signature~$\sigma$~\citep{falilusc20a}.
Thus, a program can be seen as a first-order theory over~$\sigma$.
The translation from rules to formulas is not trivial due to the need to handle all possible terms occurring in the rule.
Here, we use a simplified version of this translation, which is sufficient for the examples considered in this paper.
Formal results apply to any translation that captures the intended meaning of rules as formulas.
We identify each rule~${\mathit{Head} \leftarrow \mathit{Body}}$ with the universal closure of the formula~$B \to \mathit{Head}$, where~$B$ is the conjunction of all literals in~$\mathit{Body}$ after replacing~$\mathit{not}$ by~$\neg$.
In the resulting formulas, every variable or constant occurring as an argument of an arithmetic function is assumed to be of the sort integer, similarly to the translation by~\citet{lifschitz21a}.
We often write rules as formulas~$B \to \mathit{Head}$ omitting the reference to the universal closure.
For instance,
we identify the rule listed in Line 4 of Listing~\ref{lst:property}  with the universal closure of formula
$
q(N-1,k) \to q(N,k+1),$
where~$N$ is a variable and~$k$ is an object constant, both of the integer sort.


%

%

\emph{Interpretations} are defined as usual for many-sorted first-order languages~\cite[Section~1.2.2]{lifschitz08c}.
%
An \emph{interpretation}~$I$ of a signature~$\sigma$ is \emph{standard} if it satisfies the following conditions:
\begin{itemize}
\item its domain of the sort integer is the set of all numerals, and its domain of the sort general contains all precomputed terms,

\item all numerals and object constants of the sort general are interpreted as themselves,

\item arithmetic function are interpreted as customary in arithmetic, and

\item predicate constants corresponding to comparison are interpreted according to some fixed total order such that numerals are contiguous and ordered as~usual.
\end{itemize}
%
The value $t^I$ assigned to a ground
term~$t$ over~$\sigma$ by an interpretation~$I$ and the
 satisfaction relation (denoted, $\models$) between an interpretation of~$\sigma$ and a
sentence over~$\sigma$ are defined recursively, in the usual
way~\cite[Section~1.2.2]{lifschitz08c}.
An interpretation is called a {\em model} of a \emph{theory} -- a (possibly infinite)  set of sentences -- when it satisfies every sentence in this theory.
%
%


\mbox{\citet{peaval04b,peaval05a}} introduced the quantified logic of here\nobreakdash-and\nobreakdash-there.
One of its applications is an alternative definition of stable models for logic programs.
A key benefit of this alternative is that it bypasses a reference to a grounding process when characterizing stable models.
The many\nobreakdash-sorted case of the quantified logic of here\nobreakdash-and\nobreakdash-there was recently studied by~\citet{falite24a}. We review it here considering a signature~$\sigma$ as described above and standard interpretations.
We start by introducing some notation.
%
%
%
Let $I$ to be a standard interpretation over signature~$\sigma$.
%
If $\boldt$ is a tuple $t_1,\dots,t_n$ of ground terms, then~$\boldt^I$
is the tuple $t_1^I,\dots,t_n^I$ of values.
By~$\At{I}$ we denote the set of atoms of the form~$p(\boldt)$ such that~\hbox{$I \models p(\boldt)$}, where~$p$ is a predicate symbol and $\boldt$ is a tuple of precomputed terms.
An \emph{HT\nobreakdash-interpretation} over~$\sigma$ is a pair
$\langle \HH,I\rangle$, where
$I$ is a standard interpretation and
$\HH\subseteq\At{I}$.
(In terms of Kripke models with two worlds,~$\HH$
describes the predicates in the here-world and~$I$ captures the there-world).
The satisfaction relation~$\modelsht$ between
HT\nobreakdash-interpretation $\langle \HH, I\rangle$ of~$\sigma$
and a sentence~$F$ over~$\sigma^I$ is defined recursively:
\begin{itemize}
\item
  $\langle \HH, I\rangle \modelsht p(\boldt)$,
  if $p(\boldt^I)\in \HH$;
\item
$\langle \HH, I\rangle \modelsht t_1=t_2$ if $t_1^I=t_2^I$;
\item
$\langle \HH, I\rangle \modelsht F\land G$ if
$\langle \HH, I\rangle \modelsht F$ and
$\langle \HH, I\rangle \modelsht G$;
\item
$\langle \HH, I\rangle \modelsht F\lor G$ if
$\langle \HH, I\rangle \modelsht F$ or
$\langle \HH, I\rangle \modelsht G$;
\item
  $\langle \HH, I\rangle \modelsht F\to G$ if
    (i) $\langle \HH, I\rangle \not\modelsht F$ or $\langle \HH, I\rangle \modelsht G$,
    and
(ii)
    $I \models F\to G$;
  \item
    $\langle \HH, I\rangle\modelsht\forall X\,F(X)$ with~$X$ of the general sort,
  if $\langle \HH, I\rangle\modelsht F(t)$
    for every precomputed term~$t$;

\item
  $\langle \HH, I\rangle\modelsht\forall X\,F(X)$ with~$X$ of the integer sort,
 if $\langle \HH, I\rangle\modelsht F(\overline{n})$
  for every integer~$n$;

\item
  $\langle \HH, I\rangle\modelsht\exists X\,F(X)$ with~$X$ of the general sort,
 if $\langle \HH, I\rangle\modelsht F(t)$
  for some precomputed term~$t$;
\item
  $\langle \HH, I\rangle\modelsht\exists X\,F(X)$ with~$X$ of the integer sort,
 if $\langle \HH, I\rangle\modelsht F(\overline{n})$
  for some integer~$n$.
\end{itemize}
If $\langle \HH, I\rangle \modelsht F$ holds, we say that $\langle \HH, I\rangle$ \emph{satisfies}~$F$ and that $\langle \HH, I\rangle$ is an \emph{HT\nobreakdash-model} of~$F$.
%

It is easy to see that
$(\At{I},I)$ is an HT\nobreakdash-model of a sentence $F$ whenever $I$ is a model of~$F$.
About a model~$I$ of a theory~$\Gamma$, we say it is \emph{stable} if, for
every proper subset~$\HH$ of~$\At{I}$, HT\nobreakdash-interpretation
$\langle \HH,I \rangle$ does not satisfy~$\Gamma$.
In application to theories of a single sort (and thus in the absence of arithmetic function symbols), this definition is equivalent to the original definition by~\citet{peaval04b,peaval05a}.
In addition, if the theory is finite, then this definition of a stable model is also equivalent to the definition of such model in terms of the operator~SM \citep{feleli07a,feleli11a}, when all predicate constants are considered to be ``intensional''.
In the sequel, we sometimes abuse our terminology and notation and identify set~$\At{I}$ of ground atoms with interpretation $I$.%

\subsection{Parameterizable Subprograms}\label{sec:parametrized-collective}

We now review the key concepts introduced  by~\cite{gekakasc17a} when discussing the multi-shot framework. We then use these concepts to formally characterize the collective control described in Section~\ref{sec:motiv}.
A {\em program declaration} is an expression of the form
$\hbox{\tt \#program}\ m(p_1, \dots , p_j ),$
where $m, p_1, \dots, p_j$ are symbolic constants. We call $m$ the name of the declaration
and $p_1, \dots p_j$ its parameters/placeholders.
Different occurrences of
program declarations with the same name are assumed to share the same parameters. In this way, each name is associated with a unique
parameter specification.
A {\em \clingo-program} is a list of  rules and declarations.
Listings~\ref{lst:property} exemplifies this concept.
The {\em scope} of a program declaration in a \clingo-program consists
of the set of all rules following this declaration up to
the next program declaration or the end of the list.
In Listing~\ref{lst:property}, the scope of
the declaration in Line~1 consists of atom~{\tt q(0,0)}, while that
in Line~3 contains \mbox{\tt q(N,k+1):-(N-1,k)}.
Given \clingo-program $R$
along with a symbolic constant~$m$, we define $R(m)$ as the set of all
rules  in the scope of all occurrences of program
declarations with name $m$. We often refer to $R(m)$ as a subprogram of $R$.
All rules outside the scope of any (explicit) program declaration
are thought of being implicitly preceded by {\tt \#program base}.
Take $R_1$ to denote Listing~\ref{lst:property},
$R_1(base) = \{q(0,0)\}$
and $R_1(property) = \{q(N-1,k) \to q(N,k+1) \}$.

A {\em valuation}~$v$ of a set of placeholders~$\PH$ is a function that maps elements in this set to precomputed terms.
%
Given a  program/formula $F$, a set $\ph$ of placeholders and valuation~$v$ on~$\ph$, by $v(F)$ we denote
a program/formula constructed from $F$ by substituting each of the occurrences of some placeholder~$c \in \ph$ with the value~$v(c)$ assigned to it by the valuation~$v$.
Given a name $m$ with associated parameters/placeholders $p_1,\dots , p_j$,
and valuation~$v$ on $\{p_1,\dots , p_j\}$, we say that~$v(R(m))$ is {\em the $v$\nobreakdash-instantiation of subprogram}~$R(m)$.
For instance,
if~$v$ is the valuation~$\{k \mapsto 42\}$, then $v(R_1(property))$ consists of the universal closure of~$q(N-1,42) \to q(N,42+1)$.
We are now almost ready to formally characterize the collective control that provides specifications for instantiations of subprograms of interest and then considers these collectively as the union.
Let us define one more concept. A {\em subprogram-spec} is a triple
$
[m,\ph,v],
$
where $m$ is a name, $\ph$ is a set of associated parameters/placeholders with $m$, and~$v$ is a valuation on $\ph$.
Given (i)~\clingo-program~$R$ and (ii)~set
$[m_1,\ph_1,v_1],\dots,[m_j,\ph_j,v_j]$
of {\em subprogram-specs}, the collective control associates this pair with the logic program composed of the rules
\beq
v_1(R(m_1))\cup v_2(R(m_2))\cup\cdots\cup v_j(R(m_j)).
\eeq{eq:unionsubps}
For instance,
command line {\tt \%clingo property -c n=100} (discussed in Section~\ref{sec:motiv}) assumes \clingo-program in Listing~\ref{lst:property} and the set composed of the following subprogram-specs:
$[base,\emptyset,()],\  
[property,\{k\},(k\mapsto 0)],\  
\cdots,\ [property,\{k\},(k\mapsto 99)].$
Although  \clingo-programs provide means to group rules into subprograms, the expression~\eqref{eq:unionsubps} clearly states that the object resulting from an application of the collective control to a \clingo-program looses any aspect of the modularity.

As customary in logic programming, its programs define signatures implicitly.  Namely, predicate and object symbols occurring in a program form its signature.
If arithmetic operations are present the signature is infinite containing all numerals.

\section{Modular Logic Programs}\label{sec:modular}
Our ultimate goal is to define  parametrized  modular programs that are capable to capture
semantics of \clingo-programs with collective control declaratively
in a way that such modular programs  treat individual subprograms with reverie so that each instantiation of a subprogram will  have its individual meaning devoid of the rest of the context.

Prior to that we introduce modular logic programs that are close relatives to
 first-order modular logic program proposed by \citet{harlie16a}.
 In that work,
 first-order modular logic program was a collection of logic
programs with some predicates identified as intensional and others as extensional. Thus, some members of that collection could be
viewed as ``definitions/modules'' for concepts captured by intensional predicates.
%
%
Here, we allow arithmetic and use a concept called  intensionality statements introduced by \citet{fanlie23a} to claim more control over granularity of modules.


%

This section is organized as follows. We start by introducing simple intensionality statements -- a special case of such statements. We then define modules and modular programs that use intensionality statements, intuitively, to pin the scope of module's applicability. The section culminates in defining parametric modular programs.

\subsection{Simple Intensionality Statements}\label{sec:simple-int}
\citet{fanlie23a} introduced intensionality statements to generalize the notion of intensional predicate constants.
This concept allows some predicate constants to be intensional on some arguments while being extensional on others.
This fine granularity came with the price that checking certain interesting properties is an undecidable problem.
In particular, this is true for deciding whether the granular version of splitting theorem
studied by  \citet{fanlie23a}
is applicable.
In the sequel,
the splitting theorem  by  \citet{fanlie23a} plays an important role. 
For these reasons, we find of value to focus on a special case of intensionality statements, which
 we call {\em simple}.
Despite the fact that they are less general they are sufficient for our purposes and importantly more tractable.
Before we proceed towards the presentation of the formal definitions let us take a moment to discuss the terms {\em intensional} and {\em extensional}. This terminology roots in  deductive databases, where tuples in the database are seen as purely extensional in the sense that the whole extension of the predicate is known. Derived predicates are then purely intensional in the sense that only their definition is fixed.
Here, we adjust this terminology to allow ourselves to be more  detailed. We can point at a predicate and some of its arguments as the ones being ``defined'', while the same predicate on the remaining arguments is considered  ``known''.
This is important in our context.
Indeed, as mentioned in the introduction, one of the distinguished uses of  ${\mathit{ASP}\!+\!\mathit{control}}$ is for modeling dynamic domains. These applications form a natural example where  the same predicate is used both as intensional and extensional depending on its arguments. In a modular program describing a dynamic domain, a subprogram may naturally represent the effects of the actions at a given time step, where the same predicate is used to represent the state of the world at different time steps. In this case, the predicate is intensional for the arguments representing the current time step, and extensional for the arguments representing the previous time step.

%

A \emph{simple intensionality statement}~$\kappa$ over a signature~$\sigma$ is a function mapping each predicate symbol~$p/n$ in~$\sigma$ to a set of~$n$\nobreakdash-tuples satisfying the following conditions:
(i)~each element of the tuple is either a variable or a precomputed term of~$\sigma$, and
(ii)~no variable occurs twice in a tuple.
In the following, we name the $i$-th element of a tuple assigned by a simple intensionality statement as~$X_i$ and variations when it is a variable.
We say that a predicate symbol is (purely) \emph{extensional} when its associated set is empty, and that it is (purely) \emph{intensional} when it contains a tuple of variables (with no ground terms).
When all predicate symbols are purely intensional or purely extensional, then the simple intensionality statement corresponds to the notion of intensional predicate constants~\citep{feleli11a}.
When the arity of~$p/n$ is clear from the context, we write~$\kappa^p$ instead of~$\kappa(p/n)$.
Consider, for instance, predicate symbol~$\mathit{q}/2$ and let $\kappa^\mathit{q}$ be  set~$\{ \tuple{X,1} , \tuple{X,2} \}$.
Intuitively, this intensionality statement states that the ground atoms formed by predicate constant~$q$ with its arguments of the form~$\tuple{t,1}$ or~$\tuple{t,2}$ are intensional for any term~$t$; otherwise, these atoms are extensional.
For an intensionality statement~$\kappa$ and a predicate symbol~$p/n$, by~$\pdef_\kappa^p(\boldX)$
we denote the formula
\begin{gather}
  \bigvee_{(t_1,\dots t_n)\in \kappa^p}
  {\Big(\bigwedge_{\hbox{ {\small $\ba{c} t_i \hbox{ is not a variable}\\  (1\le i\le n)\ea$}}}{X_i=t_i}\Big)}
  \label{eq:pdef}
\end{gather}
where~$\boldX = \tuple{X_1,\dots,X_n}$ is a tuple of distinct variables of the appropriate length and sort (in the sequel we adopt this convention and use $\boldX$ to denote tuples of variables).
To each simple intensionality statement~$\kappa$, we associate a set~$\EM{\kappa}$ of sentences, called the \emph{extensional axioms} of~$\kappa$, containing a sentence of the form
\begin{gather}
  \forall\boldX \left( \neg \pdef_\kappa^p(\boldX) \to p(\boldX) \vee \neg p(\boldX) \right)
  \label{eq:ext.axiom}
\end{gather}
for every predicate symbol~$p/n$ in~$\sigma$.
For a theory~$\Gamma$ over~$\sigma$,
we say that a standard interpretation~$I$ is $\kappa$\nobreakdash-\emph{stable} if it is a stable model of~$\Gamma \cup \EM{\kappa}$.
When it is clear from the context we sometimes drop $\kappa$ as a subscript of $\pdef_\kappa^p$.

Let us illustrate these concepts on examples.
 Assume the signature $\sigma_1$ composed of a single predicate $q$.
Consider~$\kappa_1^\mathit{q} = \{ \tuple{X,1}, \tuple{X,2}  \}$; then~$\pdef_1^\mathit{q}(X_1,X_2)$ is~$(X_2 = 1 \vee X_2 = 2)$
(here and in the sequel we understand $\pdef_i$ as an abbreviation for $\pdef_{\kappa_i}$).
Thus, extensional axioms formula $\EM{\kappa_1}$ is equivalent to the the universal closure of
\begin{gather}
\big(X_2 \neq 1\wedge X_2  \neq 2\big) \rightarrow  \big(q(X_1,X_2)\vee \neg q(X_1,X_2)\big).
\label{eq:example.ext.axiom}
\end{gather}
Consider theory $\Gamma_1$ composed of the universal closures of formulas
 $q(X,0)\rightarrow q(X,1),$ and  $q(X,1)\rightarrow q(X,2).$
%
Sets
$\{ q(1,3)\} \hbox{ and } \{q(0,0), q(0,1), q(0,2)\}$
are among the $\kappa_1$-stable models of~$\Gamma_1$.
In case of the former, note that~$q/2$ is not intensional for~$\tuple{1,3}$ and hence the extensional axiom~\eqref{eq:example.ext.axiom} allows $q(1,3)$ to be either true or false.
In case of the latter, note that~$q/2$ is not intensional for~$\tuple{0,0}$ either; hence, the extensional axiom allows $q(0,0)$ to be either true or false.
Atoms~$q(0,1)$ and~$q(0,2)$ must now be true.
On the other hand, set $\{ q(0,1) \}$ is not a $\kappa_1$-stable model of~$\Gamma_1$ because~$q/2$ is intensional for~$\tuple{0,1}$ and there is no rule to derive it from.

There are some well known general properties about traditional logic programs and their stable models.
For example, we can conclude that an atom is not a member of any answer set when it does not occur as a head of any rule. A similar result is also the case for $\kappa$-stable models.
Proposition~\ref{prop:correctness} below presents this result formally.
We call a substitution~$\theta$ {\em simple} with respect to signature~$\sigma$ whenever all values assigned by $\theta$ are precomputed terms over $\sigma$ (we drop the reference to the signature when it is clear from the context).

\begin{proposition}\label{prop:correctness}
  Let~$I$ be a $\kappa$-stable model of~$\Gamma$,
  $p/n$ be predicate symbol,
  and~$\boldt$ be a tuple of terms of length~$n$
  such that~$I \models  \pdef_{\kappa}^p(\boldt)$.
It is the case that  $I \not\models p(\boldt)$ when there is no rule of the form~$B\rightarrow p(\boldr)$  in~$\Gamma$
  and simple variable substitution~$\theta$ such that~$(\boldr\theta)^I = \boldt$ and~$I \models B\theta$.
\end{proposition}

\noindent
Intuitively, condition ${I\models \pdef_{\kappa}^p(\boldt)}$ means that given a simple intensionality statement~$\kappa$, theory $\Gamma$ ``defines'' this predicate on the tuples associated with~it.

%
%

\subsection{Modules and Modular Programs}\label{sec:modprograms}
A \emph{\definition} over signature $\sigma$ is a pair
$\Module{\kappa}{\Pi},$
where
$\kappa$ is a simple intensionality statement over $\sigma$ and
$\Pi$ is a program over~$\sigma$;
%
we say that a~$\kappa$\nobreakdash-stable model of~$\Pi$ is a \emph{model} of the module. 
For instance, consider \definition~$\Delta_1$ over $\sigma_1$ to stand for
$  \Module{\kappa_1}{\Gamma_1},$ 
where~$\sigma_1$,  $\kappa_1$ and~$\Gamma_1$ are defined in Section~\ref{sec:simple-int}.
%
The models of~$\Delta_1$ are the $\kappa_1$-stable models of $\Gamma_1$. 

A \emph{modular program}~$\mp$ is a pair~$\tuple{\kappa, \mathcal{M}}$ where~$\kappa$ is a simple intensionality statement and~$\mathcal{M} = \{ \Delta_1,\dots,\Delta_n \}$ is a finite set of \definitions over~$\sigma$ such that
\begin{gather}
  \text{$\forall \boldX \left( \pdef_i^p(\boldX) \to \pdef^p(\boldX)  \right)$ is valid}
  \label{eq:modular.program.pdef.requirement}
\end{gather}
for every predicate symbol~$p/n$ in~$\sigma$ and every module~$\Delta_i$ in~$\mathcal{M}$;
where
~$\pdef^p$ and $\pdef_i^p$ respectively are the formulas associated with~$p/n$ by~$\kappa$ and each~$\kappa_i$.
Intuitively, condition~\eqref{eq:modular.program.pdef.requirement} states that every atom defined in a submodule must be intensional in the modular program.
A standard interpretation~$I$ is a \emph{stable model/answer set} of~$\mp$ if it is a model of each of its \definitions~$\Delta_i$ and it satisfies
formula
\begin{gather}
 \forall {\boldX}\big(\pdef^p(\boldX) \wedge \neg \pdef_1^p(\boldX) \wedge \dotsc \wedge \neg \pdef_n^p(\boldX) \to \neg p(\boldX)\big)
  \label{eq:modular.program.pdef.formula}
\end{gather}
for every predicate symbol~$p/n$ in~$\sigma$.
Intuitively, formula \eqref{eq:modular.program.pdef.formula} states that every intensional atom that is not defined in any submodule must be false in all stable models.

Consider a sample modular program~$\mp_1$ over $\sigma_1$ whose simple intensionality statement is defined by $\kappa^q=\tuple{X,Y}$ (in other words, predicate symbol~$q$ is purely intensional); and
set of modules consists of~$\Delta_0$, $\Delta_1$ and~$\Delta_2$, where
 \begin{align*}
\Delta_0 &= \Module{\kappa_0}{ \{ q(0,0) \} },\hbox{ and}
\\
\Delta_2 &= \Module{\kappa_2}{ \{ q(0,2)\rightarrow q(0,3),\, q(0,3)\rightarrow q(0,4) \} }
\end{align*}
with~$\kappa_0^\mathit{q} = \{ \tuple{0,0} \}$ and~$\kappa_2^\mathit{q} = \{ \tuple{X,3}, \tuple{X,4}  \}$; and $\Delta_1$ is understood as earlier.
%
The only stable model of~$\mp_1$ is
\begin{gather}
\{q(0,0),\ q(0,1),\ q(0,2),\ q(0,3),\ q(0,4)\}.
\label{eq:ex.stable.model}
\end{gather}
We assume that an empty conjunction is equivalent to~$\top$.
The condition~\eqref{eq:modular.program.pdef.formula} for~$q$ is equivalent to the universal closure of %
%
$
  (X_1 \neq 0 \wedge X_2 = 0) \vee \neg (0 \leq X_2 \leq 4) \to \neg q(X_1,X_2).
$
Hence,
every atom of the form~$q(t_1,t_2)$ is false when
both~${t_1 \neq 0}$ and~${t_2 = 0}$ or when ${t_2 \not\in \{0,1,2,3,4\}}$.
All models of~$\Delta_1$ where~$q(t_1,0)$ is false must also satisfy that~$q(t_1,1)$ and~$q(t_2,2)$ are false, and all models of~$\Delta_2$ where~$q(t_1,2)$ is false must also satisfy that~$q(t_1,3)$ and~$q(t_1,4)$ are false.
Since~$q(t_1,0)$ is false for every~$t_1 \neq 0$, then atoms~$q(t_1,t_2)$ with~$t_1 \neq 0$ and~$t_2 \in \{1,2,3,4\}$ must be false as well.
Hence, it remains to discuss atoms of the form~$q(0,t_2)$ with~$t_2 \in \{0,1,2,3,4\}$.
Atom~$q(0,0)$ must be true because  all the models of~$\Delta_0$ satisfy it.
Atoms~$q(0,1)$ and~$q(0,2)$ must be true because they are true in the only models of~$\Delta_1$ satisfying~$q(0,0)$.
Atoms~$q(0,3)$ and~$q(0,4)$ must be true because they are true in the only models of~$\Delta_2$ satisfying~$q(0,2)$.
Note that, set \eqref{eq:ex.stable.model} is exactly the unique~$\kappa$\nobreakdash-stable model of the program containing all rules in~$\Delta_0$, $\Delta_1$ and~$\Delta_2$.
This is not a coincidence, as we show in the next section.

\paragraph*{Relation between modular programs and non-modular ones.}

For every term~$t$, we define~$[t]$ recursively as follows:
\begin{itemize}
  \item if~$t$ is a variable or a precomputed term, then~$[t] = t$;
  \item if~$t$ is~$t_1 \odot t_2$ with~$\odot \in \{ +, -, \times \}$ and both~$t_1$ and~$t_2$ are ground terms, then~$[t]$ is the numeral~$\overline{n_1 \odot n_2}$ with~$\overline{n_1} = [t_1]$ and~$\overline{n_2} = [t_2]$;
  \item if~$t$ is~$t_1 \odot t_2$ with~$\odot \in \{ +, -, \times \}$ and at least one of~$t_1$ and~$t_2$ is a variable, then~$[t]$ is the expression~$[t_1] \odot [t_2]$;
\end{itemize}
For example, $[0+1] = 1$, and~$[X+1] = X + 1$.
If~$\boldt = (t_1,\dots,t_n)$ is a tuple of terms, then~$[\boldt]$ denotes the tuple~$([t_1],\dots,[t_n])$.

A \definition $\Module{\kappa}{\Pi}$ 
is called \emph{simple} when for every atom~$p(\boldt)$ in the head of a rule in~$\Pi$, there is a tuple~$\boldu$ in~$\kappa^p$ and a substitution~$\theta$ such that~$[\boldt] = \boldu\theta$.
A modular program is called \emph{simple} if all its \definitions are simple.
Modular program $\mp_1$ defined above is simple.
%
Indeed, we can check that \definition~$\Delta_1=\Module{\kappa_1}{\Gamma_1}$ 
is simple.
We have two atoms in the heads of its rules, namely, $q(X,1)$ and $q(X,2)$.
A substitution that maps $X_1$ to $X$ will turn tuples in~$\kappa_1$ into the arguments of these atoms.
Similarly, we can check that \definitions~$\Delta_0$ and~$\Delta_2$ are also simple. 
As another example,
let us consider \definition
\begin{gather}
\tuple{ \{\tuple{X,1}\}, \{q(N-1,0) \to q(N,0+1) \}}
\label{ex:theta0}
\end{gather}
which corresponds to the program~$\mathit{property}(k)$ in Listing~\ref{lst:property} with~$k=0$.
This \definition is also simple.
Take the substitution that maps~$X$ to~$N$ and note that~$[{0}+{1}]$ is numeral~$1$.

We next introduce the notion of \emph{coherent modular program} and show that every coherent modular program has the same stable models as a non-modular program obtained by taking the union of all rules in its \definitions.
To do so, we need to introduce the notion of a \emph{dependency graph} of a modular program.
%
%
Given a modular program~$\mp = \tuple{\kappa, \{ \Delta_1,\dots,\Delta_n  \}}$, the \emph{(directed) graph of dependencies},
denoted~$\G(\mp)$, is  defined as follows:
\begin{itemize}
  \item Its vertices are pairs~$(p,i)$ with~$p$ a predicate symbol and~$1 \leq i \leq n$.

  \item It has an edge from~$(p,i)$ to~$(q,j)$ when for some rule~$r$ of~$\mp$,
  \begin{itemize}
  \item
  there is an atom of the form $p(\boldt)$ in the head of~$r$, and

  \item
  there is a nonnegated atom of the form of~$q(\boldt')$ in the body of~$r$,  and

  \item
  there are tuples~$\boldu_i$ in~$\kappa_i^p$ and~$\boldu_j$ in~$\kappa_j^q$, and~substitutions~$\theta$ and~$\theta'$ such that ${[\boldt] = \boldu_i\theta}$ and~${[\boldt'] = \boldu_j\theta'}$.
\end{itemize}
\end{itemize}
We say that a strongly connected component of~$\G(\mp)$ is \emph{contained in \definition~$\Delta_i$} if all vertices in this component are of the form~$(p,i)$.

As an example dependency graph~$\G(\mp_1)$ contains three vertices $(q,0)$, $(q,1)$ and~$(q,2)$ and an edges from $(q,2)$ to $(q,1)$ and from $(q,1)$ to $(q,0)$.
It is easy to see that
the strongly connected component $(q,i)$ is contained in \definition $\Delta_i$ for any~$i \in \{0,1,2\}$.

A simple modular program~$\mp$ is called \emph{coherent} if it satisfies the following two conditions:
\begin{itemize}
  \item every strongly connected component of the dependency graph~$\G(\mp)$ is contained in some \definition; and

  \item every pair of \definitions~$\Module{\kappa_i}{\Pi_i}$ and~$\Module{\kappa_j}{\Pi_j}$ in~$\mp$ satisfy that, for every predicate symbol~$p$, there is no pair of tuples~$\boldu_i \in \kappa_i^p$ and~$\boldu_j \in \kappa_j^p$ with~$i \neq j$, such that~$\boldu_i$ and~$\boldu_j$ unify. %
  Two tuples $\boldt_i=\langle t^i_1,\dots t^i_m \rangle$ and~$\boldt_j=\langle t^j_1,\dots t^j_m \rangle$ of terms \emph{unify} if there is a substitution~$\theta$ of the variables that applied to both~$\boldt_i$ and~$\boldt_j$ yields $t^i_k=t^j_k$, $1\leq k\leq m$
\end{itemize}

\begin{theorem}\label{thm:union.modular.program}
  The answer sets of a coherent modular program~$\mp$ are the same as the answer sets of the program obtained by taking the union of all rules in its \definitions.
\end{theorem}

\begin{theorem}\label{thm:modular.programs.complexity}
  Deciding if a modular program is coherent is feasible in polynomial time.
\end{theorem}

\subsection{Parametric Modular Logic Programs}\label{sec:modlog}
As illustrated in Section~\ref{sec:parametrized-collective},
in practice, modules are often defined in a parametric way, that is,  in terms of some parameters that can be instantiated in different ways to obtain different modules.
To capture this idea, we introduce the notion of \emph{parametric \definition}s that generalize \definitions\ introduced in Section~\ref{sec:modprograms}.

Consider a set $\ph$ of object constants not occurring in $\sigma$,
then by $\sigma^\ph$ we denote the signature $\sigma$ extended with the object constants in $\ph$.
By $\ph^+$ we denote the signature composed of the object constants in $\ph$, numerals, and arithmetic function constants.

A \emph{parametric intensionality statement} over  $\sigma^\ph$ is a
function mapping each predicate symbol $p/n$ in $\sigma$ to a set of $n$-tuples satisfying the conditions:
\begin{itemize}
\item each element of the tuple is either a variable,
a precomputed term,
 or a ground term over signature~$\ph^+$;
\item  no variable occurs twice in a tuple.
\end{itemize}
A \emph{parametric \definition} over  $\sigma$ is a triple
$\trdef{\ph}{\chi}{\Pi},$
 where
 $\chi$ is a parametric intensionality statement over $\sigma^\ph$ and
 $\Pi$ is a formula (program understood as a set of formulas) over~$\sigma^\ph$.
 In the sequel, to illustrate relevant concepts we use the parametric \definition
 \beq
\tuple{\{k\}, \chi_1, \{q(N-1,k) \to q(N,k+1) \}},
\eeq{eq:d2}
where  we assume the signature composed of a single predicate $q$
and parametric intensionality statement $\chi^q_1$ is characterized by the singleton set consisting of \tuple{X,k+1}.

Given a parametric intensionality statement~$\chi$ over $\sigma^\ph$ and substitution~$\Theta$ on $\ph$ by
$\chi\Theta$ we denote a function mapping  each predicate symbol $p/n$ in $\sigma$ to the following set of $n$-tuples
$
\{{\tuple{[t_1\Theta],\dots,[t_n\Theta]}}\mid \tuple{t_1,\dots,t_n}\in \chi^p\}.
$
%
For a simple substitution $\Theta$ on $\ph$,
$\chi\Theta$ forms a simple intensionality statement. (We defined simple substitutions in Section~\ref{sec:simple-int}.)

For example, let~$\Theta_1$ be  a simple substitution that maps $k$ to precomputed term $1$. Then
$$
\begin{array}{lll}
(q(N-1,k) \to q(N,k+1))\Theta_1&\hbox{\;\;\;\;   is \;\;\;\;\; }&q(N-1,1) \to q(N,1+1);\\
\chi_1^q\Theta_1&\hbox{\;\;\;\;      is \;\;\;\;\; }&\{\tuple{X,2}\}.\\
\end{array}
$$
Given  a parametric \definition  $\Phi=\trdef{\ph}{\chi}{\Pi}$ over signature $\sigma$
and a simple substitution $\Theta$ on~$\ph$, we can construct a \definition
 $\Phi_\Theta=\tuple{\chi\Theta,\Pi\Theta}$ (as introduced in
Section~\ref{sec:modprograms}). In the scope of this section it is convenient to call such \definitions\
 {\em instances}.
The {\em signature of any instance} of the parametric module is that of a parametric module, namely,~$\sigma$.
When the set of placeholders $\ph$ is empty in a considered parametric module, it serves a role of an instance itself.

For example,  let~$D^k$
 denote parametric \definition in~\eqref{eq:d2}.
Take $\Theta_0$ be a simple substitution that maps $k$ to precomputed term $0$.
Then, instance $D^k_{\Theta_0}$ coincides with  \definition~\eqref{ex:theta0}.
Similarly, if~$\Theta_1$ is a simple substitution that maps $k$ to precomputed term $1$, then
instance~$D^k_{\Theta_1}$ stands for
$\tuple{\{\tuple{X,2}\}, \{q(N-1,1) \to q(N,1+1) \}}.$

We are  ready to look into the example discussed in
Section~\ref{sec:motiv}, namely, \property parametric-program (Listing~\ref{lst:property}) that comes with collective control~\eqref{eq:A}.
We identify subprogram~{\tt property(k)} with parametric \definition~$D^k$ ($k\geq 0$) introduced above. 
Substitutions $\Theta_i$ ($0\leq i\leq n$)  map parameter $k$ into precomputed term (integer) $i$.
We identify subprogram~$base$ in the running example with a parametric \definition
\beq
 \tuple{\emptyset,\{\tuple{0,0}\}, \{q(0,0) \}}
\eeq{eq:dprime}
that we denote as $D$ and which coincides with the \definition $\Delta_0$ introduced in Section~\ref{sec:modprograms}
(this module is an instance itself as its set of placeholders is empty).
We view the  \property parametric-program together with the collective control as modular program
\beq
\tuple{\kappa,\{ D, D^k_{\theta_0},\dots,D^k_{\theta_n}\}},
\eeq{eq:mpproperty}
where the signature of this modular program contains predicate $q/2$ and
the simple intentionality statement $\kappa$ defines $q$ as purely intensional.
This modular program has a unique answer set, which, for example, in case when $n=99$  coincides with the answer set of program~\eqref{lst:unfolded}.
This last observation is not a coincidence and is supported by Theorem~\ref{thm:union.modular.program}. In fact, that theorem gives us formal grounds for our earlier actions of identifying subprograms $base$ and~{\tt property(k)} with parametric \definitions~$D$ and $D^k$ ($k\geq 0$), respectively.

 \section{Utility of Declarative Approach to Modularity}

We now illustrate how this declarative approach to modularity allows us to make formal claims about a program with control without a reference to underlying workings of \clingo. Rather, these arguments  naturally accompany intuitive readings of the parametric programs that we supply \clingo with.

 Consider the \property program in Listing~\ref{lst:property}. Our first intuition is that
 its {\tt base} subprogram states that
 \vspace{-5pt}
 \begin{equation}
\hbox{ pair $\tuple{0,0}$ of integers  has property $q$. }\label{eq:claimbase}
 \end{equation}
In Section~\ref{sec:modlog} we argued that  the
{\tt base} subprogram can be identified with  \definition~$D$ defined in~\eqref{eq:dprime}.
Any model of this instance supports claim~\eqref{eq:claimbase} formally.

 Our second intuition is that the {\tt property(k)} subprogram states that given $i$ is an integer substituted for $k$ within this subprogram and property $q$ holds for some pair~$\tuple{j-1,i}$, where $j$ is an integer, then property $q$ also holds for the pair~$\tuple{j,i+1}$.
Take $i$ to denote an arbitrary integer and $\Theta_i$ to denote a simple substitution that maps $k$ into $i$.
Recall that we view the {\tt property(k)} subprogram as
parametric \definition~$D^k$; whereas  $D^k_{\Theta_i}$ captures an instance of that program obtained by replacing integer~$i$ for parameter~$k$.
With that it is easy to see that any model of this instance supports the intuition we claim.

Last but not least is grasping the overall meaning of the \property program when control is in place and formally reasoning about its properties.
First, the control specifies which exact instances of the subprograms are being part of an overall program being processed.
Second, it implicitly defines the signature under which the program is interpreted.
To continue with our running example, consider control~\eqref{eq:A}.
This control together with the considered parameterizable subprograms warrants the signature of a single predicate $q/2$ and infinite set of numerals.
We now construct an inductive proof that the modular program listed in~\eqref{eq:mpproperty} (the modular program that we identify with the \property program with the collective control)
  has a unique answer set $I$ whose set $\At{I}$ of true atoms has the form
  $\{q(0,0)\;\;\ q(1,1)\;\; \ q(2,2)\;\;\ \dots\;\;\ q(n,n)\}$.
  The uniqueness of this answer set follows immediately from the following claims

  {\em Claim 1: for any natural number $i$ that is less or equal to $n$,  $q(i,i)$ belongs to $\At{I}$};

  {\em  Claim 2: for any natural number $i$ that is greater than $n$,  $q(i,i)$ does not belong to~$\At{I}$};

  {\em  Claim 3: for any distinct natural numbers $i$ and $j$,  $q(i,j)$ does not belong to~$\At{I}$}.
\\
We prove \emph{Claim~1} by induction.
 Base case. Take $i$ to be $0$, $q(0,0)$ is part of $\At{I}$ due to the observation we made as we discussed the first intuition.
 Our inductive hypothesis is that the claim in question is the case for some integer $j$ that is less than $n$.
 We now show that it is the case for $j+1$.
By inductive hypothesis $q(j,j)$ is part of $\At{I}$.  Then, $q(j+1,j+1)$ is part of $\At{I}$ due to the presence of the instance $D^k_{\theta_j}$  in~\eqref{eq:mpproperty} and
  observation we made as we discussed the second intuition.

For \emph{Claim~2},
condition~\eqref{eq:modular.program.pdef.formula} of the definition of stable models of modular programs for the case program~\eqref{eq:mpproperty} and predicate symbol~$q/2$ has the~form
\beq
\forall X_1 X_2\big(\top\wedge (X_1\neq 0\vee X_2\neq 0)\wedge X_2\neq 1\wedge\dots\wedge X_2\neq n\rar\neg q(X_1,X_2)\big).
\eeq{eq:touse}
This implies that whenever the second argument, let us call it $m$, of predicate $q$ is greater than $n$, atom of the form $q(\cdot,m)$ is not part of $\At{A}$. Claim 2 follows from this observation. 

To show that \emph{Claim~3} holds, it is sufficient to consider the case when
$j$ is less or equal to~$n$.
Indeed, recall the argument supporting \emph{Claim~2}.
Now we prove this claim by induction on~$j$.
\emph{Base case}.
Take $j=0$. And consider arbitrary integer $i$ different from~$j$.
Trivially, $i\neq 0$. The consequence of~\eqref{eq:touse} is satisfied by $I$.
Then,
$q(i,j)$ is not part of~$\At{I}$.
\emph{Inductive step}.
Consider $0\leq j< n$.
The \emph{inductive hypothesis} states:
{\em for any natural number $i$ that is different from $j$,  $q(i,j)$ does not belong to~$\At{I}$}.
We now show that {\em for any natural number $i$ that is different from $j+1$,  $q(i,\numeral{j+1})$
does not belong to~$\At{I}$}.

Consider any natural number $i$ different from $j+1$.
Module~$D^k_{\theta_{j}}$ that has the form
$
\tuple{\{\tuple{X,\overline{j+1}}\}, \{q(N-1, j) \to q(N,j+1) \}}.
$
Formula $\lambda^q(i,\overline{j+1})$ is~$X_2=\overline{j+1}$.
Thus, ${I\models \lambda^q(i,\overline{j+1})}$.
Take $\Theta$ to denote the substitution ${N\mapsto i}$.
This is the only substitution which turns atom $q(N,j+1)$ that occurs in the head of the only rule in  $D^k_{\theta_{j}}$ into ground atom $q(i,j+1)$. Ground atom $q(N-1, j)\Theta$ forms the body of that rule and has the form $q(i-1,j)$. Given that $i\neq j+1$, it follows $i-1\neq j+1$. By inductive hypothesis  $q(i,\numeral{j+1})$
does not belong to~$\At{I}$.
By Proposition~\ref{prop:correctness},
$q(i,\overline{j+1})$ is not part of~$\At{I}$.

Note how all our formal claims about the shape and the uniqueness of the answer set of the \property program with collective control are void from the reference to the inner workings of \clingo. There is no reference to concatenation, grounding, and solving.

\section{Conclusions}
This paper champions a need for investigating a declarative approach to modularity in ASP. We take a step in this direction by introducing parametric modular programs and their declarative semantics. In the concluding section we illustrate the utility of this approach by reasoning about a sample ASP program with collective control in a formal way that is void from the reference to the inner workings of ASP systems.
This is an important step towards refining our understanding about modular formalisms and providing ASP practitioners with formal tools to reason about the properties of modular programs frequently required in the design of complex systems. Yet, we only scratched the surface of this important topic. There are many directions for future research. For example, it is important to investigate how the proposed approach can be extended to capture other forms of controls.
In practice, it is common that controls  use so called external atoms to ``deactivate'' certain parts of the subprograms ``on demand''. We believe that our framework can be extended to capture this kind of control as well, and we plan to explore this direction.
Another viable direction for future work is a development of a declarative meta-language for the
 instantiation of parametric modules. The specifications in that language would  define which instantiations of parametric modules  form the actual intended program.

\paragraph{Acknowledgements.}
We would like to thank the anonymous reviewers for their valuable feedback that allowed us to improve the presentation of several points.
This work was supported by
the National Science Foundation CAREER award 2338635, USA,
and the DFG grant SCHA 550/15, Germany.
Any opinions, findings, and conclusions or recommendations expressed in this material are those of the authors and do not necessarily reflect the views of the National Science Foundation.

\bibliographystyle{plainnat} 
\bibliography{krr,local,procs}

\end{document}